\RequirePackage{fix-cm}
\let\oldvec\vec
\documentclass[twocolumn]{svjour3}    
\let\vec\oldvec
\smartqed  
\usepackage{graphicx}
\usepackage{amsmath}
\usepackage{amsfonts}
\usepackage{amssymb}
\usepackage[numbers]{natbib}
\usepackage[T1]{fontenc}
\usepackage[utf8]{inputenc}
\usepackage{float}
\restylefloat{table}
\usepackage[table]{xcolor}
\usepackage{soul}

\begin{document}


 \title{Deep Learning Architecture for Automatic Essay Scoring}


\author{Tsegaye Misikir Tashu    \and Chandresh Kumar Maurya      \and Tom\'a\v{s} Horv\'ath 
}

\authorrunning{Tsegaye M.Tashu et al.} 

\institute{Tsegaye Misikir Tashu \and Chandresh Kumar Maurya \and
        Tom\'a\v{s} Horv\'ath \at
              ELTE E\"otv\"os Lor\'and University, Faculty of Informatics,\\ Department of Data Science and Engineering, \\Telekom Innovation Laboratories\\ P\'azm\'any P\'eter s\'et\'any 1/C, 1117, Budapest, Hungary \\
              \email{misikir@inf.elte.hu}           
          \and
            Tom\'a\v{s} Horv\'ath \at
              Pavol Jozef \v{S}af\'arik University in Ko\v{s}ice, Faculty of Science, Institute of Computer Science\\Jesenn\'a 5, 040 01 Ko\v{s}ice, Slovakia\\
              \email{tomas.horvath@inf.elte.hu}
             \and 
             Tsegaye Misikir Tashu \at
             Wollo University, Kombolcha Institute of Technology, College of Informatics,\\ 208 Kombolcha, Ethiopia
}

\date{Received: date / Accepted: date}

\maketitle

\begin{abstract}
Automatic evaluation of essay (AES) and also called automatic essay scoring has become a severe problem due to the rise of online learning and evaluation platforms such as Coursera, Udemy, Khan academy, and so on. Researchers have recently proposed many techniques for automatic evaluation. However, many of these techniques use hand-crafted features and thus are limited from the feature representation point of view. Deep learning has emerged as a new paradigm in machine learning which can exploit the vast data and identify the features useful for essay evaluation. To this end, we propose a novel architecture based on recurrent networks (RNN) and convolution neural network (CNN). In the proposed architecture, the multichannel convolutional layer learns and captures the contextual features of the word n-gram from the word embedding vectors and the essential semantic concepts to form the feature vector at essay level using max-pooling operation. A variant of RNN called Bi-gated recurrent unit (BGRU) is used to access both previous and subsequent contextual representations.   The experiment was carried out on eight data sets available on Kaggle for the task of AES. The experimental results show that our proposed system achieves significantly higher grading accuracy than other deep learning-based AES systems and also other state-of-the-art AES systems.

\end{abstract}

\section{Introduction}
\label{intro}

\label{sec:introduction}

Automatic essay scoring (AES) system was introduced to alleviate the workload of the assessor and to improve the feedback cycle in the teaching-learning process. Since its introduction, a number of research activities have been carried out \cite{Page1966,Attali2011,ZUPANC2017118,tseg2018,Tashu2019}. The task of AES was regarded as a machine learning problem that learns to approximate the assessment process using handcrafted features with supervised learning \cite{Attali2006,Foltz1999,Deerwester1999}. Some of the features extracted most frequently are essay length, sentence length, grammar correctness \cite{Burstein}, readability \cite{zesch-etal-2015} and textual coherence \cite{Chen2013}. These handcrafted features, however, require much human commitment and usually require a complex implementation for each new feature. 

Recently, natural language processing (NLP) has achieved some success in switching from such linear models via sparse and handcrafted feature inputs to nonlinear neural network models via dense inputs \cite{Goldberg2016}. The recent success comes from the application of deep learning models on NLP tasks where these models are capable of modelling intricate patterns in data without handcrafted features. Since these methods do not depend on manually produced features, they can also be used to solve problems in an end-to-end fashion. SENNA \cite{Collobert2011} and neural machine translation \cite{Bahdanau2015NeuralMT} are two remarkable examples in NLP that work without external task-specific knowledge. These successes in the processing of natural language tasks attracted scientists in the field of AES  \cite{Alikaniotis2016,taghipour,dong-zhang-2016}. The most commonly used model in AES \cite{Alikaniotis2016,taghipour}, among others, is a recurrent neural network model called long short term memory (LSTM).

Scoring essays utilizing handcrafted features like, for example, essay length, sentence length, grammar correctness or readability face the following problems: First, it might be used by students as an option to cheat the system by writing and submitting a well-structured essay which is off-topic. Every well-written essay that does not address the question topic may receive a good score from an AES system because of linguistic features such as text structure and surface. If every essay submitted to an AES system is evaluated as a standard essay input, then it may degrade user confidence in the AES engine \cite{higgins2006}. Second, creating these handcrafted features is tedious, time-consuming and sometimes inefficient. Therefore, new approaches that score essay semantically without handcrafted features have to be used.

To semantically score essays, latent semantic models such as Latent Semantic Analysis (LSA) \cite{Deerwester1999,Shermis2003,Shermis2013},  Generalized Latent Semantic Analysis (PLSA) \cite{Islam2010}, Latent Dirichlet Allocation (LDA), Bilingual Topic Model (BLTM) \cite{Kakkonen2006}, Word Mover's Distance (WMD) \cite{tseg2018,Tashu2019} and Content Vector Analysis (CVA) \cite{Attali2006,Attali2011} were proposed and applied.

{The success of deep learning in the areas of natural language processing and the problems related to manual feature engineering led to the use of neural network methods for Automatic essay scoring. The works of Taghipour and Ng} \cite{taghipour}; Cummins and Rei \cite{Cummins:2018}; Wang et al. \cite{wang-etal-2018};  Jin et al. \cite{jin-etal-2018-tdnn};  Farag et al.\cite{farag-etal-2018-neural};  Zhang and Litman \cite{zhang-litman-2018-co}  {used mostly long short term memory (LSTM) based neural network models for AES. Most of these studies used the combination of neural models and hand-crafted features, since their objective was mostly to score essays using discourse structure and coherence attributes. Thus, our work looks at advancing AES by exploring other architectures that can capture more in-depth and essential contextual information in scoring essays semantically. }

In this study, we introduce an extended latent semantic model utilizing the advantages of deep learning. Our model, called DeLAES, addresses the above stated problems by modifying and combining existing latent semantic models. DeLAES uses a multichannel CNN with different window sizes with the Max Pooling operation and also bidirectional gated recurrent neural networks to capture more in-depth and essential contextual information.

Instead of using input representation based on bag-of-words, DeLAES considers an essay as a sequence of words with rich contextual structure, and it retains maximum contextual information in its projected latent semantic representation. The model first projects each word in its context onto a low-dimensional continuous feature vector using skip-gram models. Then, it directly learns and captures the contextual features at the word n-gram level using a three-channel convolutional neural network. Second, the model does not discover and aggregate all word n-gram characteristics, but only the essential semantic concepts, to form a feature vector at the essay level using max-pooling. Third, the essay-level feature vector of each channel is then passed on to a bidirectional gated recurrent neural network that performs a nonlinear transformation to extract high-level semantic information from the word order of each channel. The extracted semantic high-level representation from each channel is then aggregated to obtain the final semantic representation at the essay level and fed into a fully connected layer for score prediction using the Sigmoid function.
The contribution of this paper can be summarized as follows:
\begin{enumerate}
    \item A new latent semantic model is introduced that captures the previous as well as subsequent contextual features for contextual structures at both the word n-gram and essay levels using multichannel convolutional pooling operations with bidirectional gated recurrent neural networks;
   
    \item An experiment is performed on the data set provided as part of the automated essay scoring competition on the Kaggle website that contains essays by students from eight different prompts (essay discussion questions) or topics;
    
    \item Experimental results are compared to published state-of-the-art results and also with baseline approaches based on deep learning methods.
\end{enumerate}

The rest of the paper is organized into six sections. Section \textbf{2} describes related works that are relevant to our research. Section \textbf{3} presents the proposed DeLAES Architecture,  Section \textbf{4} presents the overall experimental settings,  implementation and evaluation of the proposed system and Section \textbf{5} the results of our system DeLAES and other baselines. Section \textbf{6} presents the discussion and conclusions.

\section{AES Related works}
\label{sec:related}
The research on automatically evaluating and scoring essays is ongoing for more than a decade where Machine Learning (ML), NLP and artificial neural networks (NN) were used for evaluating essay question answers. In this section, we present the characteristics of the majority of AES systems developed by commercial organizations and also AES systems introduced by the academic community. 

Project Essay Grade (PEG) was the first AES system developed by Ellis Page and his colleagues \cite{Page1966}. It evaluates and scores essays by measuring trins and proxes. A trin is defined as an intrinsic higher-level variable, such as punctuation, fluency, diction, grammar, the number of paragraphs, average sentence length, the length of an essay in words, counts of other textual units, etc., which as such cannot be measured directly and has to be approximated by means of other measures, called proxes. \cite{Shermis2002,BoWang2005,ZhangLi2014}. However, the exact set of textual features underlying each dimension and the details concerning the derivation of the overall score were not disclosed \cite{Ben-Simon2007,Shermis2002}. The system uses regression methods to score new essays based on a training set.

Intelligent Essay Assessor (IEA) uses on LSA \cite{Foltz1999} to infer characteristics that define the content, organizational, and development-related attributes of the essay. IEA also extracts other essential features that measure lexical sophistication, grammatical, mechanical, stylistic and organizational aspects of essays using natural language processing techniques.

The system uses different attributes to measure the above aspects in essays, such that, context (e.g. semantic similarity), lexical sophistication , grammar (grammatical errors and grammatical error types), mechanics (e.g. spelling, upper and lower case), style, organization and development (e.g. sentence/sentence coherence, general essay coherence and topic development). IEA requires training essay set with a representative sample of a manually annotated essay by experts. 

IntelliMetric was designed and introduced in 1991 by Vantage Learning as proprietary AES \cite{Shermis2003}. IntelliMetric analyzes the semantic, syntactic and discursive attributes of essays to form a composite sense of meaning. The attributes used in IntelliMetric can be grouped into two main groups (content and structure). The content attributes are used to evaluate the subject covered, the breadth of content, the support of advanced concepts, the logic of discourse, the cohesion and consistency of the purpose and main idea. The structural attributes evaluate spelling, punctuation, letter cases, grammar, syntactic literacy, syntactic diversity, sentence complexity, use, legibility, and subject-binding. The system uses multiple predictions with linear analysis, Bayesian approach and LSA in predicting the final score, and combines the models into a single final score. 

E-Rater \cite{Attali2006} works by extracting features such as grammatical errors, word usage errors, writing mechanics errors, presence of essay-based discourse elements, development of essay-based discourse elements, style flaws, content vector analysis (CVA) to evaluate current word usage, an alternative, differentiated measurement of word usage, based on the relative frequency of a word in high-scoring versus low-scoring essays. Such an attribute that takes into account the correct use of prepositions and collocations, and diversity in the formation of sentence structures and finally apply regression modelling to predict the score.

The Lexile Writing Analyzer is part of the Lexile Framework for Writing \cite{Smith2009} developed by MetaMetrics. The system is the score, genre, prompt and punctuation independent and uses the Lexile Writer measure which is an estimate of the student's ability to express language in writing based on factors related to semantic complexity (the level of words used) and syntactic sophistication (how words are written in sentences). The system uses some attributes that represent approximate values for writeability. Lexile perceives the ability to write as a fundamental individual characteristic. 

SAGrader \cite{brent2006} is a proprietary AES system developed  By IdeaWorks, Inc. SAGrader combines a number of linguistic, statistical and artificial intelligence approaches to evaluate the essay automatically. The operation of the SAGrader is as follows: The instructor first specifies a task in a prompt. Then the teacher creates a section in which he identifies the "desired properties" vital elements of knowledge (facts) to be included in a right answer, as well as the relationships between these elements via a semantic network. Fuzzy logic allows the program to recognize the features in students' essays and compare them with the desired ones. Finally, an expert system evaluates student essays based on the similarities between the desired and observed characteristics. Students immediately receive feedback by reporting their results, along with detailed comments on what they have done well and what remains to be done. The system provides both results and meaningful feedback through ontology-based information extraction.

The system that uses logical reasoning to recognize errors in a statement in an essay was proposed by Gutierrez et al. in \cite{Gutierrez:2012}. The system first transforms the text into a set of relevant clauses using the Open Information Extraction methodology and integrates them into the domain ontology using a manually selected vocabulary mapping. The system determines whether these statements contradict ontology and, thus, domain knowledge. This method regards falsity as inconsistency with respect to the domain. 

The automated scoring engine called CRASE \cite{Lottridge2013}, developed by Pacific Metrics, passes through the following three phases of the scoring process: Identification of inappropriate attempts, attribute extraction and scoring. The attribute extraction step is constructed around six aspects of the essay: ideas, typesetting, organization, voice, wording, conventions and written presentation. The system analyzes a sample of student responses that have already been scored to create a model of grader scoring behaviour. It is a Java-based application that runs as a web service. The system is adaptable to the configurations used to build machine learning models and to mix human and machine. The application also generates text-based and numerical feedback that can be used to improve the essays.

The first evaluation system made publicly available was Rudner's Bayesian Essay Test Scoring system called BETSY \cite{Rudner2002}. BETSY uses Naive Bayes models to classify texts into different classes (e.g. Pass or Fail) based on content (e.g. Word Uni-Gram and bi-gram) and style attributes. The classification is based on the assumption that each attribute is independent of another. BETSY has only proven to be a demonstration tool for Bayesian approach to essay evaluation.

An automated evaluation engine with compiled and publicly available source code called LightSIDE was introduced in 2010 by Mayfield and Rose \cite{Mayfield:2010}. LightSIDE was developed as a tool for non-experts to effectively use text mining technology for a variety of purposes, including essay evaluation. It allows the selection of the set of attributes and algorithms to build a prediction model. The set of attributes focuses mainly on n-grams, part of speech tags, and "counting" attributes and also the system have a feature that allows the user to enter the code for new attributes manually.

As the existing AES techniques which are using LSA do not consider the word sequence of sentences in the documents and the creation of word by document matrix is somewhat arbitrary. Chali and Hasan \cite{chali2013} proposed an AES system that calculates the syntactic similarity between two sentences by analyzing the corresponding sentences in syntactic trees and measuring the similarity between the trees. The so-called shallow semantic tree kernel method allows parts of semantic trees to be synchronized. The kernel function returns the similarity value between a pair of sentences based on their semantic structures. 

Islam and Hoque \cite{Islam2010} developed an AES system using generalized latent semantic analysis (GLSA), which develops a system that uses grams for document matrix instead of word for document matrix, as in LSA. The system uses the following steps in the grading procedure: Pre-processing of training essays, removal of stop words, word origin, selection of n-gram index terms, n-gram when creating document matrices, calculation of singular value decomposition (SVD) from n-gram to document matrix, dimensional reduction of SVD matrices and calculation of similarity evaluation. The main advantage of GLSA is adherence to the word order in sentences. In order to reduce memory and time consumption without lowering the performance of automated scoring in comparison with human scoring, \cite{ZhangLi2014} proposed incremental singular value decomposition as a part of incremental LSA to score essays when the dataset is massive.

An extension of existing AES systems was introduced in \cite{ZUPANC2017118} by incorporating additional semantic coherence and consistency attributes. They designed coherence attributes by transforming sequential parts of an essay into the semantic space and measuring changes between them to estimate coherence of the text and consistency attributes that detect semantic errors using information extraction and logic reasoning. The resulting system, named ``sage-semantic automated grader for essays'', provides semantic feedback for the writer and achieves significantly higher grading accuracy compared with other state-of-the-art AES systems. 

An investigation on the effectiveness of using semantic vector representations for the task of AES was performed in \cite{jin2017}. According to the evaluation results on the standard English dataset, the effectiveness brought by the proposed semantic representations of essays depends on the learning algorithms and the evaluation metrics used. On the other hand, the effectiveness of individual semantic features is stable with respect to different numbers of dimensions. 

In \cite{Fauzi}, an AES system based on n-gram and cosine similarity were described. N-Gram was used for feature extraction and modified to split by word instead of by letter so that the word order would be considered. Based on evaluation results, this system got the best correlation of 0.66 by using unigram on questions that do not consider the order of words in the answer. For questions that consider the order of the words in the answer, bi-gram has the best correlation value, by 0.67. 

The architecture of an AES system based on a rubric, which combines automated scoring with human scoring, was proposed in \cite{Yamamoto2018}. The proposed rubric has five evaluation viewpoints: contents, structure, evidence, style, and skill as well as 25 evaluation items which are subdivided in viewpoint. At first, the system automatically scores 11 items included in the style and skill such as sentence style, syntax, usage, readability, lexical richness, and so on. Then it predicts scores of style and skill from these items' scores by multiple regression models. It also predicts contents' score by the cosine similarity between topics and descriptions

AES systems that are introduced above are based on regression methods applied to a set of carefully handcrafted designed features. The process of feature engineering is the most challenging part of building AES systems. Moreover, it is challenging for humans to consider all the factors that are involved in assigning a score to an essay. To alleviate these issue, AES systems that learn features and relation between an essay and its score automatically were introduced using deep neural networks\cite{Alikaniotis2016,taghipour,dong-zhang-2016}. 

{Recently, neural network models have been introduced into AES, making the development of handcrafted features unnecessary or at least optional. Alikaniotis et al.} \cite{Alikaniotis2016} and Taghipour and Ng \cite{taghipour} presented AES models that used Long Short-Term Memory (LSTM) networks. Differently, Dong and Zhang \cite{dong-zhang-2016} used a Convolutional Neural Network (CNN) model for essay scoring by applying two CNN layers on both the word level and then sentence level. Later, Dong et al. \cite{dong-etal-2017} presented another work that uses attention pooling to replace the mean over time pooling after the convolutional layer in both word level and sentence levels. Dong and Zhang \cite{dong-zhang-2016} {showed that a two-layer Convolutional Neural Network (CNN) outperformed other baselines (e.g., Bayesian Linear Ridge Regression) on both in-domain and domain adaptation experiments on the ASAP dataset. This model was later improved by employing attention layers. Specifically, the model learns text representation with LSTMs which can model the coherence and co-reference among sequences of words and sentences, and uses attention pooling to capture more relevant words and sentences that contribute to the final quality of essays} \cite{dong-etal-2017}. 

Song et al. \cite{song-etal-2017} {proposed a deep model using gated recurrent unit  and bidirectional gated recurrent unit for identifying discourse modes in an essay. Chen and He} \cite{chen-he-2014} {studied the usefulness of prompt independent text features and achieved a human machine rating agreement slightly lower than the use of all text features for prompt-dependent essay scoring prediction. A constrained multitask pairwise preference learning approach was proposed by Cummins et al.} \cite{Cummins:2018}  {by combining essays from multiple prompts for training. However, as shown by Dong and Zhang} \cite{dong-zhang-2016} and Zesch et al. \cite{zesch-etal-2015} {straightforward applications of existing AES methods for prompt-independent AES lead to a poor performance. In the same year, Jin et al.} \cite{jin-etal-2018-tdnn} {proposed a two-stage deep neural network (TDNN) learning framework to address a poor permanence of the approach proposed by Cummins et al.} \cite{Cummins:2018} {The TDNN utilizes the prompt-independent features to generate pseudo training data for the target prompt, on which a bidirectional LSTM was used to learn a rating model consuming semantic, part-of-speech, and syntactic signals. Their TDNN model outperforms the baselines, and leads to promising improvement in the human-machine agreement.}

{Most of the proposed neural network models aim in scoring essays using discourse structure, coherence, part-of-speech, and semantics attributes and the results were promising, but more research has to be conducted to have a more accurate and robust AES system.  Thus, our work looks at advancing AES by exploring other architectures that will improve the predictive accuracy by capturing more in-depth and essential contextual information in scoring essays semantically. }

In this paper, we propose a novel deep learning architecture for AES. The proposed architecture is enhanced by bidirectional GRU (BGRU) using a multichannel convolutional layer with Max-Pooling operation, referred to as Max-Pooling based BGRU using multichannel convolutional layer. The basic idea of the new architecture is based on the following considerations. The one-dimensional convolutional filters with different window size in the convolutional layer perform in extracting n-gram features at different positions of the essay and reduce the dimensions of the essay. Since all the n-gram features extracted by the convolutional layer are not relevant to the semantics of the essay, Max-Pooling operation is used to select the most essential n-gram features. The BGRU is used to learn and extract the preceding and succeeding contextual features from the outputs of the Max-Pooling layer. The outputs from each BGRU channel will be then aggregated using the fully connected layer, and the final score will be predicted using the sigmoid layer.

\section{Max-Pooling based Bidirectional Gated Recurrent Network using Multi-channel Convolutional Filters Model}

The prediction task addressed in this paper can be briefly described as follows:
Given a set of labeled essays $X=\{(\mathbf{x},y)\:|\: \mathbf{x}=(w_n)_{n\in\mathbb{N}}, y\in[0,1]\}$, where each essay $\mathbf{x}$, a sequence of words, is assigned a score $y$, the tasks is to learn a scoring function $s:(w_n)_{n\in\mathbb{N}}\rightarrow[0,1]$ such that the difference between the true score $y$ and the predicted score $\hat{y}=s(\mathbf{x})$ for any essay $\mathbf{x}$ is minimal.

\subsection{DeLAES Architecture}
\begin{figure*}
    \centering
 \includegraphics[width=0.6\textwidth]
 {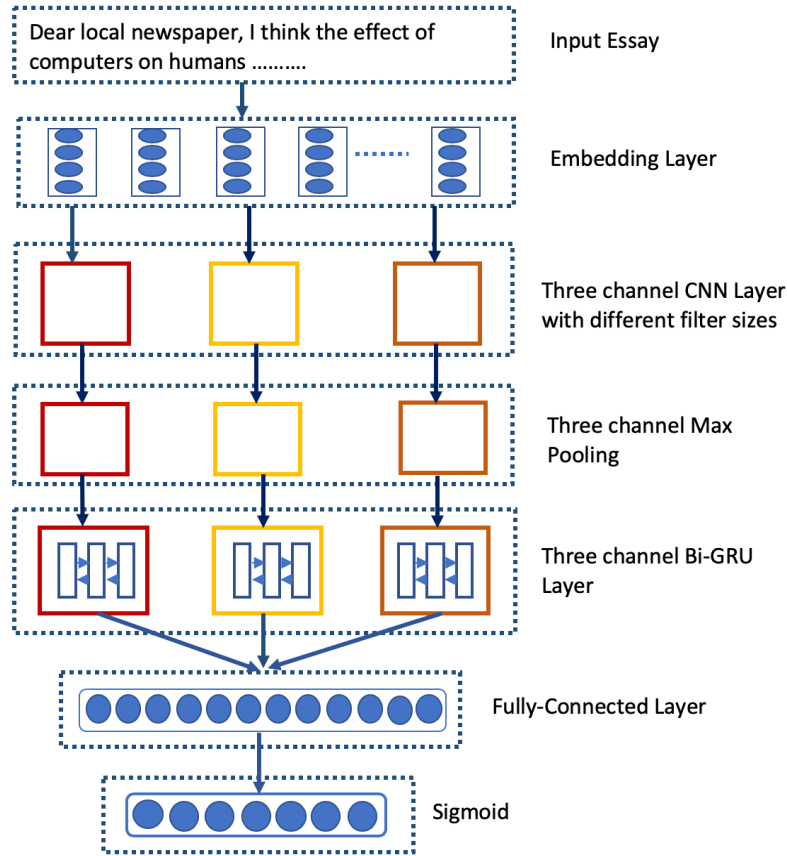}
 \caption{The proposed DeLAES architecture.}
  \label{fig:architecture}
  
\end{figure*}

The proposed architecture is illustrated in Figure \ref{fig:architecture} which can be summarized as follows (detailed descriptions of its layers will be introduced in the following subsections): 
\begin{enumerate}
    \item The embedding layer maps the words 
    of an input essay 
    into latent vectors utilizing pre-trained word vectors.

    \item Learning parameters, weights and biases are initialized with random values (they will be adjusted through training).
    
    \item A three channel convolutional layer is employed that extracts contextual features for each word with  its  neighboring  words  defined  by  a  window size (we used three different window sizes); input the dense matrix representation of the essay to the convolutional layer along with the initialized parameters. The convolutional layer then produces n-gram level features.
    
    \item A max-pooling operation is applied that accepts the n-gram level feature representation and discovers salient word-n-gram features to form an essay-level feature vector. 
    
    \item The BGRU layer obtains the preceding and the succeeding contextual features. 
    This layer extracts the final high-level semantic  feature vector representation for each input word sequence essay.
    
    \item The fully connected payer combines the semantic feature vector representation of each word to obtain a comprehensive semantic feature representation; 
    
    \item The comprehensive semantic feature representation of the essay is fed into the sigmoid function to get the predicted labels;
    
    \item The loss of the prediction is computed using  mean squared error.
    
    \item An optimization algorithm is used for loss minimization, adjusting the weights and biases based on the computed loss. For this study, the  RMSPro [12] optimizer was used. 
    
    \item The above described process is repeated until the model reaches the desired or the highest accuracy possible. Afterwards, the trained model can be used for evaluation of new essays by predicting the corresponding scores. 
\end{enumerate}

 In the next subsections, each layer of our model is described in detail.

\subsubsection{Text Representation Layer}
The first layer of our model projects each word $w$ into a $d$ dimensional latent space, resulting in a vector $v(w)\in\mathbb{R}^d$. Given an input essay $\mathbf{x} = (w_1, w_2, \dots, w_m)$ of a sequence of $m$ words, its corresponding word embedding results in a matrix $E\in\mathbb{R}^{d\times m}$ with columns $v(w_1)$, $v(w_2)$, $\dots$, $v(w_m)$. 
The resulting word embedding matrix $E$ will be used as an input for 1D convolution for feature extraction 
and will be learned during training. The most popular word embedding model proposed by Mikolov \cite{Mikolov:2013}, trained using the skip-gram method by maximizing the average log probability of all words, is used in this work. 

\subsubsection{CNN Layer}
CNNs, originally proposed for computer vision, have  shown strong performance on natural language processing \cite{Collobert2011} and text classification tasks \cite{wang2015,Zhang2015}. 

Once the dense representation $E$ of the input sequence $\mathbf{x}$ is calculated using the embedding layer, it is fed into the recurrent layer of the network. However, it might be advantageous for the network to extract local (salient) features from the sequence before feeding it into the recurrent operation. This can be performed by applying a convolution layer on the output of the embedding layer. 
 
The convolution operation can be considered as a sliding window based feature extraction. It is designed to capture the word n-gram contextual features.

A one dimensional (1D) convolution layer of width $k$ works by moving a sliding window of size $k$ over the word embedding matrix ($E$) representation of each essay \cite{kim2014} and applies the filter to each window in the sequence of vectors\footnote{The {i}th, {(i+1)}th, \dots, {(i+k-1)}th columns of $E$, respectively.} $(E_{\star,i}$, $E_{\star,i+1}$, $\dots$, $E_{\star,i+k-1})$ for $1\leq i\leq m-k+1$,
This results in a concatenated dense vector $c_i$ which is $Conv(\mathbf{e}_i)$ as defined in equation \ref{equ:conv}.
\begin{equation}
\label{equ:conv}
    c_i = g(W.\mathbf{e}_i + b)
\end{equation}

for each $\mathbf{e}_i= (E_{\star,i}, E_{\star,i+1}, \dots, E_{\star,i+k-1}) \in \mathbb{R}^{d\times k}$, for each $1\leq i\leq m-k+1$. 
where $g$ is a non-linear activation function that is applied element-wise, $W$ is the feature transformation function and $b$ is the bias of the network. We used rectified linear unit (ReLU) as the nonlinear activation function because it can improve the learning dynamics of the network and significantly reduces the number of iterations required for convergence in deep networks.  At the convolutional layer, words within their contexts are projected to vectors that are close to each other if they are semantically similar. The output of the convolutional layer is a sequence of  feature vectors whose length is proportional to the length of the input word sequence.

This filter is applied to each possible window of words $\mathbf{e}_{1}$, $\mathbf{e}_{2}$, $\dots$, $\mathbf{e}_{m-k+1}$ in the essay

to produce a feature map
\begin{equation}
  \mathbf{c}=(c_1, c_2, \dots , c_{m-k+1}) \in \mathbb{R}^{m-k+1}
\end{equation}

\subsubsection{Max-pooling layer}
A  sequence of local contextual feature vectors are extracted at the convolutional layer, one for each word n-gram. These local features need to be aggregated to obtain an essay-level feature  vector representation. Since we want to extract words that have significant importance in the semantics of the essay, we apply a max-pooling operation on the convolution contextual feature representation.  
The max pooling operation obtains the most useful local features produced by the convolutional layers, i.e.,  selects the highest neuron activation value across all local feature vectors. The $m$ vectors are then extracted using a max pooling layer, resulting in a single dimensional vector $\mathbf{p}$ as defined in equation \eqref{max-pooling}.
 \begin{equation}
\label{max-pooling}
  \mathbf{p}_j = \max_{1<i \leq{m}}c_i[j]
\end{equation}
$\mathbf{p}_j$ denotes $j^{th}$ element of the max pooling layer $\mathbf{p}$, $c_i$ is the $i^{th}$ element of the local feature vector $\mathbf{c}$. The effect of the max-pooling operation is to get the most salient information across all window positions. 

What we have described is the process by which one feature is extracted from one filter window size. Our proposed model uses 100 filters with different window sizes to obtain multiple features, and these features are then passed into a bidirectional gated recurrent unit for further processing. Therefore, the convolution layer can be seen as a function that extracts important feature vectors from n-grams. Since this layer provides n-gram level information to the subsequent layers of the neural network, it can potentially capture local contextual dependencies in the essay and consequently improve the performance of the model.

\subsubsection{Bidirectional Gated Recurrent Unit (BiGRU)}

The sequence of word embeddings learned by using the embedding layer  and further processed by convolution layer and max-pooling operation is then passed to a BGRU network layer.  The gated recurrent unit (GRU) was recently introduced as an alternative to the LSTM to make each recurrent unit to adaptively capture dependencies of different time scales \cite{Chung:2014}.  Similarly to the LSTM unit, the GRU has gating units that modulate the flow of information inside the unit, however, without having a separate
memory cells. Below are the updates performed at each $t \in \{ 1,....,T\}$ in a bidirectional GRU parameterized by weight matrices $W_z, W_r , W_h,  U_r, U_z , U_h, U_o$.

\noindent Forward updates:
\begin{equation}
    \overrightarrow{Z_t} = sigmoid\big(\overrightarrow{W_z} X_t + \overrightarrow{U_z} h_{t-1}\big)
\end{equation}
\begin{equation}
   \overrightarrow{r_t} =sigmoid\big(\overrightarrow{W_r} X_t + \overrightarrow{U_r} h_{t-1}\big)
\end{equation}
\begin{equation}
    \overrightarrow{\hat{h_t}}= tanh\big(\overrightarrow{W_h} X_t + \overrightarrow{U_h} (r_t \odot h_{t-1})\big)
\end{equation}

\begin{equation}
    \overrightarrow{h_t} = (1- \overrightarrow{Z_t}) \odot \overrightarrow{h_{t-1}} + \overrightarrow{Z_t} \odot \overrightarrow{\hat{h_{t}}}
\end{equation}

\noindent Backward updates:
\begin{equation}
    \overleftarrow{Z_t} = sigmoid\big(\overleftarrow{W_z} X_t + \overleftarrow{U_z} h_{t-1}\big)
\end{equation}
\begin{equation}
   \overleftarrow{r_t} =sigmoid\big(\overleftarrow{W_r} X_t + \overleftarrow{U_r} h_{t-1}\big)
\end{equation}
\begin{equation}
    \overleftarrow{\hat{h}_t}= tanh\big(\overleftarrow{W_h} X_t + \overleftarrow{U_h}(r_t \odot h_{t-1})\big)
\end{equation}
\begin{equation}
    \overleftarrow{h_t}=(1- \overleftarrow{Z_t}) \odot \overleftarrow{h_{t-1}} + \overleftarrow{Z_t} \odot \overleftarrow{\hat{h}_{t}}
\end{equation}

 where $\odot$ is an element-wise multiplication.
The activation $h_t$ at time $t$ is a linear interpolation between the previous activation $h_{t-1}$
and the candidate activation $\hat{h_t}$. An update gate $Z_t$
decides how much the unit updates its activation or content. The reset gate $(r_t)$ is used to control access to the previous state $h_{t-1}$ and compute a proposed update $\hat{h_t}$. When off ($r_t$ close to $0$), the reset gate effectively makes the unit act as if it is reading the first symbol of an input sequence, allowing it to forget the previously computed state.

\subsubsection{Bidirectional Models}
The idea of bidirectional models (BRNN, BLSTM and BGRU) is
to present each training input sequence forwards and backwards to two separate networks, both of which are connected to the same output layer.  This means that for every point in a given sequence, the bidirectional model has complete, sequential information about all points before and after it. At each time stamp, the concatenation of the forward and backward hidden state produces the hidden state of the bidirectional model. The prediction at time stamp $t$ is computed by combining the outputs of forward model $\overrightarrow{h_t}$ and the backward $\overleftarrow{h_t}$, formulated as follows:
\begin{equation}
    H_t=[ \overrightarrow{h_t},\overleftarrow{h_t}]
\end{equation}
The bidirectional model have shown improved results in sequence learning tasks and speech processing \cite{Jinmiao1999,LIU2019}.
\subsubsection{Fully-connected Hidden Layer}
The outputs that are obtained from each hidden state are concatenated together to form the neural feature vector representation for the essay, and we pass the concatenated feature vector into the final sigmoid layer for score prediction.  
\begin{equation}
    y_{out}= sigmoid\big(W_f.H +bf\big)
\end{equation}
where $H$ is the concatenation from the outputs of the recurrent layer, $H = (h_1,h_2,\dots,h_i)$, $W_y$ and $b_y$ are the parameters of the linear layer and $y_{out} \in [0,1]$. The output of the layer is a normalized score of the essay, and we also normalize all gold-standard scores to [0,1] and use them to train the network and  we re-scaled the output of the network to the original score range and use the re-scaled scores to evaluate the performance of the method.

\subsubsection{Objective and Optimization}
We use the RMSProp optimization algorithm \cite{Dauphin2015} to minimize the mean squared error (MSE) loss function over the training  data, defined as
\begin{equation}
\label{mse}
  MSE(y,\hat{y}) = \sum\limits_{i=1}^{K} (y_i,\hat{y_i})^2
\end{equation}
where $y$ is the gold standard score and
$\hat{y}$ is the model predicted score. Given $K$ training essays and their corresponding normalized gold standard scores $y_i$ (the actual scores), the predicted normalized score by the model is $\hat{y_i}$ for all training essay and then the network parameters are updated such that the MSE is minimized. %

\section{Experimental Settings}
In this section, we elucidate the empirical setup, data sets used, performance metrics, baselines, and experimental evaluation.

\subsection{Parameters}
To determine the hyperparameters and to avoid overfitting, we divide the training data into two sets that do not overlap, called training and validation datasets, respectively. In the experiment, the models were first trained on the training data using 10-fold cross validation, and then training parameters are optimized on the validation data. The weights of the neural network are initialized randomly, as suggested in \cite{Dauphin2015}. 
The model is trained using mini-batch based stochastic gradient descent. Each mini-batch consists of 128 training samples. To create mini-batches for training, we pad all essays in a mini-batch using a dummy token to make them have the same length. To eliminate the effect of padding tokens during training, we mask them to prevent the network from miscalculating the gradients.

\begin{table}[t]
    \centering
    \caption{The best performing hyperparameters used in NNs, grid search was used to choose the best parameters.}
    \begin{tabular}{|rc|}
        \hline
         Hyperparameters& CNN-Bi-GRU  \\
         \hline
         CNN Window Size & 2,3,4 \\
         Number of CNN Filters & 100 \\
         Batch Size & 128 \\
         Number of BGRU hidden units & 128 \\
         Dropout rate & 0.4 \\
         Number of Epochs & 40 \\
         Learning Rate & 0.001 \\
         Word Embedding dimensions & 300 \\
         \hline
    \end{tabular}
    \label{tab:parameteres}
\end{table}

\subsection{Data sets and Evaluation Methodology}
We perform the experiments on the dataset that was provided within the Automated Essay Scoring competition on the Kaggle \footnote{\texttt{https://www.kaggle.com/c/asap-aes}} website. The datasets contain student essays for eight different prompts (essay discussion questions). The anonymized students were from the USA and were drawn from three different grade levels: 7, 8, and 10 (aged 12, 13, and 15, respectively). Four datasets included essays of traditional writing genres such that persuasive, expository or narrative. The other four datasets were source-based, i.e. the students had to discuss questions referring to a previously read source document. At least two human expert graders score each training set. The authors of the datasets already divided them into fixed training and test sets. Since the test labels are not released, we split the training set into training and test sets to build scoring models and measure prediction accuracy, respectively. The characteristics of the used datasets are shown in Table \ref{tab:dataset}.

\begin{table}[ht!]
\caption{Main characteristics of datasets used in the experiment.}
\label{tab:dataset}
\centering
\begin{tabular}{|cccc|}
  \hline
  Essay\_set (ES) & \#Essay & Average\_length & Scores  \\
  \hline
  ES1 & 1783 & 350 & 2-4 \\
  ES2 & 1800 & 350 & 1-6 \\
  ES3 & 1726 & 150& 0-3 \\
  ES4 & 1772 & 130  & 0-3\\
  ES5 & 1805 & 150 & 0-4 \\
  ES6 & 1800 & 150 & 0-4 \\
  ES7 & 1569 & 250 & 0-30 \\
  ES8 & 723 & 650 & 0-60  \\
  \hline
\end{tabular}
\label{tab:example2} \centering
\label{datasets}
\end{table}

We use 10-fold cross-validation to evaluate the proposed method, since the test set used in the competition is not publicly available. In each fold,  70\% of the data is used as our training set, 10\% as the validation set, and 20\% as the test set. We train all models for 40 epochs and select the best model based on the performance on the evaluation set. The evaluation is conducted in a prompt-specific fashion(essay topic).  We also use dropout regularization to avoid overfitting \cite{Srivastava2014}. The input of the dropout is the output of the recurrent layer. Dropout randomly selects some entries in its input with probability p and resets them by assigning zero to their value. The rest of the entries will be copied to the output.

We tokenize the essays using the NLTK \footnote{http://www.nltk.org} tokenizer, lowercase the text, and normalize the gold-standard scores to the range of $[0,1]$. 
To learn the representation of each essay, the freely available word2vec \footnote{https://code.google.com/archive/p/word2vec/} word embedding was used, with an embedding for $3$ a million words/phrases from Google News trained using the approach in \cite{Mikolov:2013}.  During testing, we re-scale the system-generated normalized scores to the original range of scores and measure the performance using quadratic Weighted Kappa scores described in the following subsection.

\subsection{Evaluation Metric}

Model validation in AES systems depends on comparing the similarity between the predicted score of the model with the score given by the human raters \cite{Attali2011}. In this scenario, the scores from human judges are considered as gold standard and function as an explicit criterion for evaluating the performance of AES models. We use the most widely used evaluation metric called quadratic weighted kappa (QWK). QWK is an error metric that measures the degree of agreement between the automated scores and the resolved human scores, and is an analogy to the correlation coefficient. This metrics scores ranges from 0 to 1. In case that there is less agreement between the graders than expected by chance, this metric may go below 0.

Quadratic weighted kappa is calculated as follows. First, a weight matrix $W$ is constructed using equation \ref{weight}.

\begin{equation}
\label{weight}
         W_{i,j} = \frac{(i-j)^2}{(N-1)^2}
\end{equation}

where $i$ and $j$ are the actual rating (assigned by a human annotator or by the assessor) and the predicted rating (assigned by an AES system), respectively, and $N$ is the number of possible ratings.

An observed score matrix $O$ is calculated such that $O_{i,j}$ denotes the number of essays that receive a rating $i$ by the human annotator and a rating $j$ by the AES system. An expected score (rating) matrix $E$ is calculated as
the outer product of histogram vectors of the two (actual and predicted) ratings, assuming that there is no correlation between rating scores. The matrix $E$ is then normalized such that the sum of elements in $E$ and the sum of elements in $O$ are the same. Finally, given the matrices $O$
and $E$, the QWK score is calculated as indicated in equation \ref{QWk}.

\begin{equation}
\label{QWk}
            k= 1-  \frac{\sum\limits_{i,j} W_{i,j} O_{i,j}}{\sum\limits_{i,j} W_{i,j} E_{i,j}}
\end{equation}

\subsection{Baselines}

To analyze the potential benefits of the proposed DeLAES, we compared the predictive accuracy of our system with the following other five different versions of deep learning AES system, three classical machine learning systems and also to other state-of-the-art AES systems:

\begin{enumerate}

   \item Hybrid Variants: Experiments with three hybrid recurrent NN approaches, namely, BRNN with multichannel CNN (MCom\_BRNN), GRU with multichannel CNN (MCom\_GRU), BLSTM with multichannel CNN(MCom\_BLSTM), were carried out. The same parameter setting was used while implementing these baselines, as indicated in table \ref{tab:parameteres} whenever required.  

    \item  BRNN//BLSTM: We also used two variants of bidirectional recurrent neural network (BRNN), and bidirectional long short term memory (BLSTM) network in our experiments. We use the same parameter setting for all baseline models as indicated in table \ref{tab:parameteres} where the value of the parameter is needed. 
    
    \item { Machine learning approaches: We also experimented with the well-known machine learning approaches to support vector machine (SVM), Random Forest (RF) and Naive Bayes(NB). The default configurations provided by scikit-learn} \footnote{https://scikit-learn.org/stable/index.html} for each approach were used. 
   
   \item Finally, we also compared the experimental results of our proposed model with other state-of-the-art commercial and academic AES systems such as SAGE \cite{ZUPANC2017118} ,PEG	\cite{Page1966} , e-rater \cite{Attali2006} ,IntelliMetric \cite{Shermis2003} ,CRASE \cite{Lottridge2013} ,LightSIDE \cite{Mayfield:2010} ,IEA	-\cite{Foltz1999} and Lexile \cite{Smith2009}. Further details can be found at section \ref{sec:soa-baselines}.
\end{enumerate}

\section{Results}


\subsection{Accuracy of the Semantic-based DeLAES system}

 {In the following experiments, we compared our system with three different systems using multichannel CNN, two bidirectional neural network systems and three machine learning systems to evaluate if semantic attributes learned using both multichannel CNN and bidirectional approaches yield to better model performance.} Table \ref{base_results}  {shows the quadratic weighted Kappas for DeLAES, three other versions of recurrent neural network approaches using multichannel CNN, three bidirectional recurrent neural networks and also three classic machine learning approaches on all eight datasets. We also calculated the p-values as shown in table} \ref{tab:pvalue}  {between DeLAES and all approaches on the all the dataset by running the models using 10-fold cross validation.}

\begin{table*}[h!]

\caption{Comparison of the DeLAES system with other Deep learning based AESs. The table shows quadratic weighted Kappas, achieved on different datasets.}
\label{base_results}
\centering

\begin {tabular} {|cccccccccc|}
\hline

Model&			ES1&	ES2&	ES3&	ES4&	ES5&	ES6&	ES7&	ES8&	AVG   \\
\hline
DeLAES &		\textbf{0.927}&	 \textbf{0.932}&\textbf{0.884}&	0.870&	0.925&	\textbf{0.923}&	
\textbf{0.887}&	\textbf{0.873}&	 \textbf{0.903} \\
MCoM\_GRU&		0.884&	0.900&	0.868&	\textbf{0.900} &	0.897&	0.86&	0.826&	0.855&	0.874 \\
MCoM\_BLSTM&	0.920&	0.923&	0.856&	0.836&	\textbf{0.927}&	0.911&	0.872&	0.813&	0.882 \\

BLSTM&			0.818&	0.854&	0.789&	0.816&	0.853&	0.894&	0.811&	0.727&	0.820 \\
MCoM\_BRNN&		0.785&	0.843&	0.822&	0.834&	0.891&	0.859&	0.874&	0.820&	0.841 \\

BRNN&			0.793&	0.840&	0.737&	0.814&	0.818&	0.839&	0.837&	0.770&	0.806 \\
SVM&			0.445&	0.413&	0.448&	0.491&	0.402&	0.420&	0.438&	0.394&	0.431 \\
RF&				0.736&	0.686&	0.727&	0.716&	0.685&	0.697&	0.709&	0.726&	0.710 \\
NB&				0.637&	0.594&	0.633&	0.639&	0.594&	0.600&	0.642&	0.652&	0.624 \\

\hline
\end{tabular}
\end{table*}

The results in table \ref{tab:pvalue}  {show that the prediction accuracy was improved on six (6) out of eight (8) datasets and the predictions were significant (p-values < 0.05) on five data sets when the prediction accuracy is compared to MCom\_GRU and MCom\_BLSTM.  As most of the prediction results are statistically significant when DeLAES is compared to all other baselines, we indicate the insignificant results using \textit{*} in table} \ref{tab:pvalue}.

\begin{table*}[ht!]
\caption{The p-values resulting from the Wilcoxon signed-rank test between the $QWK$ results of the proposed DeLAES and the baselines. }
\label{tab:pvalue}
\centering

\begin {tabular} {|ccccccccc|}
\hline
Model&					ES1&	ES2&	ES3&	ES4&	ES5&	ES6&	ES7&	ES8 \\
\hline

DeLAES vs MCoM\_GRU &	0.0226&  0.0744*&  0.5751*&  0.3328*&  0.0227&  0.0050&  0.0468&  0.0166 \\

DeLAES vs MCoM\_BLSTM & 0.0093&  0.0093&  0.0166&  0.3862*&  0.0468&  0.2845*&  0.0218&  0.1394* \\
\rowcolor{gray}
DeLAES vs BLSTM & 0.0298&  0.5750*&  0.0386&  0.0298&  0.0359&  0.0202&  0.0512*&  0.0076 \\

DeLAES vs MCoM\_BRNN & 0.0125&  0.0050&  0.0218&  0.0050&  0.0166&  0.0050&  0.1141*&  0.0093 \\
\rowcolor{gray}
DeLAES vs RNN & 0.0284&  0.0125&  0.0125&  0.0298&  0.0069&  0.0366&  0.5750*&  0.0125 \\

DeLAES vs SVM  & 0.0050&  0.0050&  0.0050&  0.0050&  0.0050&  0.0050&  0.0050&  0.0050 \\

DeLAES vs RF &	 0.0069&  0.0050&  0.0093&  0.0050&  0.0069&  0.0069&  0.0069&  0.0593* \\

DeLAES vs NB & 0.0050&  0.0050&  0.0050&  0.0050&  0.0050&  0.0069&  0.0050&  0.0166 \\
\hline
\end{tabular}
\end{table*}

 {We aimed to determine whether the semantic-based essay evaluation using multichannel CNN and bidirectional models contributes to the higher prediction accuracy.  Thus, we compared the DeLAES system with the system without multichannel CNN. Therefore, the p-values evaluating these comparisons shown also in Table} \ref{tab:pvalue}   {are colored in gray. The results show that using multichannel CNN and bidirectional models leads to higher Kappa values in all eight observed datasets for the two compared system pairs (DeLAES-BRNN and DeLAES-BLSTM) and the results are also statistically significant in six of eight datasets. This proves that learning the feature representation and feature selection using multichannel CNN helps the lower layers of our model to learn the semantic feature representation of the essay in better way and improves the model prediction accuracy significantly.}

{We also compared the performance of DeLAES with three classical machine learning approaches (support vector machine (SVM), Random Forest (RF) and Naive Bayes (NB)). The results also show that the prediction accuracy was significantly (p-values < 0.05) improved in all of the datasets. In general, according to the results presented in table} \ref{base_results}, {all neural network model variants were able to learn the task correctly and work competitively compared to the baselines. However, DeLAES performs better than all other systems, outperforming MCoM-GRU by a wide margin (2.9\%) and MCoM-BLSTM by 2.0\%. The experimental results all show that the ensemble models were also performing better than the non-ensemble and lead to improvements in the prediction accuracy.}

{To verify how the proposed DeLAES systems performed,  we provided one example where DeLAES performs better and another example where DeLAES gives a worse prediction.  In example}\ref{exam:1}, {we present an essay where our DeLAES system predicts the score of the essay with identification number $9$ of set $1$ from the test set as $11$ while the actual score is $9$. }

\begin{quote}
\begin{example}
\label{exam:1}
    \textit{: `` Dear newspaper, @caps1 having kids wasting there whole lives by being on the computer to much. i sure wouldn't want my son to be like that. being on the computer too much can make you unhealthy from not exercising, not be able to enjoy nature, and keep you away from you friends and family. not much good can came out of being on the computer @num1. please newspaper take my word. @caps4 will make you well known @caps3 you agree with me on this. first of all, being on the computer all day, everyday can make you unhealthy by not exercising. \dots''}
\end{example}

\end{quote}

\begin{example}
User-defined numbered environment
\end{example}

Example \ref{tab:example2} { is an example to showcase the perfection of our DeLAES system, where an essay with identification number $101$ of set $1$ from the test was predicted accurately with a score of $10$.  This is one of the most correctly predicted examples from the test set.}

\begin{quote}
\begin{example}
   \textit{: `` dear @caps1 of the @caps2 @caps3, @caps4 teens are online @num1 chatting with friends and surfing the web. i personally think we should get off the computer. if you want to talk to someone, just call them! with a computer, it can start a lot of drama too! and lastly, we should enjoy nature! so obviouly computers have a larg effect on people! keep reading to hear my reasoning. the first reason why i think computers effect us is because we get attached to talking to people from a computer! we are use to going on facebook and @caps5 everyone. \dots''} 
   
 \end{example}
   
\end{quote}

\subsection{Comparison with the other AES Systems}\label{sec:soa-baselines}

\begin {table*}[ht!]
\caption{Comparison of the proposed multichannel CNN max-pooling operation with bidirectional gated recurrent NN based AES system (DeLAES) with other state-of-the-art systems. The table shows Quadratic Weighted Kappas, achieved on different datasets.}
\label{tab:results1}
\centering
\begin {tabular} {|cccccccccc|}
\hline
System	&ES1	&ES2		&ES3	&ES4 &ES5	&ES6	&ES7	&ES8 & AVG\\	
\hline
DeLAES &		0.93 &	 \textbf{0.93} &\textbf{0.88}& \textbf{0.87} &	0.92&	\textbf{0.92}&	\textbf{0.89}&	\textbf{0.87} & \textbf{0.90}\\
SAGE \cite{ZUPANC2017118}			& 0.93		&0.79			&0.83	&0.81 &0.89	&0.79	&0.88	&0.81 & 0.84\\	
PEG	\cite{Page1966}			&0.82		&0.72			&0.75	&0.82 &0.83	&0.81	&0.84	&0.73 & 0.79 \\	
e-rater \cite{Attali2006}         &0.82		&0.74			&0.72	&0.80 &0.81	&0.75	&0.81	&0.70 &  0.77\\		
IntelliMetric \cite{Shermis2003}	&0.78		&0.70			&0.73	&0.79 &0.83	&0.76	&0.81	&0.68 & 0.76\\	
CRASE \cite{Lottridge2013}			&0.76		&0.72			&0.73	&0.76 &0.78	&0.78	&0.80	&0.68 &  0.75\\
LightSIDE \cite{Mayfield:2010} 		&0.79		&0.70			&0.74	&0.81 &0.81	&0.76	&0.77	&0.65& 0.75 \\	
IEA	-\cite{Foltz1999}			&0.79		&0.70			&0.65	&0.74 &0.80	&0.75	&0.77	&0.69 &  0.74\\		
Lexile \cite{Smith2009}			&0.66		&0.62			&0.65	&0.67 &0.64	&0.65	&0.58	&0.63  & 0.64\\
\hline

\end{tabular}
\end{table*}

The results of the proposed DeLAES system was also compared with the state-of-the-art systems  like SAGE, PEG, e-rater, IntelliMetric, CRASE, LightSIDE, IEA, Lexile 
writing analyzer and with other state of the art AES systems where we obtained the results from  previous studies of Shermis and 
Hamner \cite{Shermis2013} and Zupanic \cite{ZUPANC2017118}. Since not all the systems are available for public experimenting, their results were obtained from the papers \cite{Shermis2013} and \cite{ZUPANC2017118}. The results of comparison to these systems are presented in Table \ref{tab:results1}.   

According to the results presented in table \ref{tab:results1}, 
our system achieves a better result on  $7$ out of $8$ dataset $(ES2,ES3, ES4,ES5, ES6, ES7$ and $ES8)$. On the remaining one dataset (ES1),  the accuracy of DeLAES was the same as SAGE. On the average (the rightmost column), DeLAES achieved much better results than other systems.



\section{Discussion and Conclusion }

In DeLAES, the purpose of using the convolutional layer is to pre-process the input text data. Due to its ability to detect local correlations of spatial or temporal structures, the convolution layer is ideally suited to extract n-gram features at different positions of the text from the word vectors through the convolution filters. In text classification, the procedures for generating word embedding vectors can affect classification accuracy. The max-pooling operation is mainly used to identify the influential n-grams from each the sentence in the essay. { compared to directional models, bidirectional models can access both the preceding and succeeding contextual information. Hence, bidirectional models can be more effective in learning the context of each word in the text, as indicated in our experimental results in table} \ref{tab:results1}  . 
{Furthermore, the combination of CNN based approaches and bidirectional recurrent neural based approaches makes the understanding of text semantics more simple and improves the prediction accuracy, as indicated in our experimental results on AES datasets}.

Experiments show that the convolutional layer with the max-pooling operation and BGRU have an essential influence on the performance. It is worth to note that the lower layers (recurrent neural network) have more significant effects on the prediction accuracy than the convolutional layer, since the convolutional layer is used as n-gram based feature selection. For the convolutional layer, the convolution window size and the max-pooling operation also affect the performance, according to our experimental results. 

In text classification, in general, and specifically in AES, the methods to generate word embedding vectors can also affect the classification (score prediction) accuracy. Compared to pre-trained embedding word vectors, training random embedding word vectors requires more parameters, a huge dataset, and it causes relatively lower classification accuracy in limited iterations. The experiments show that pre-trained embedding word vectors can achieve better results than random word embedding vectors. Hence, to generate pre-trained embedding word vectors is more suitable.

All experiment results show that the combination of the convolutional layer, max-pooling operation and BGRU remarkably improves the AES system prediction accuracy. For most of the benchmark datasets, DeLAES was able to obtain significantly better results than other baseline models. It shows that DeLAES has a better classification ability and performs better than other state-of-the-art deep NNs.

In the task of AES, feature extraction and the design of classifiers are very important. LSTM and GRU have shown better performance on many real-world and benchmark text classification problems. However, it is still challenging to understand the semantics and the classification accuracy still needs to be improved. In order to solve these problems, an improved GRU method, namely DeLAES, was presented in this paper utilizing multichannel convolutional layer,  max-pooling operation and BGRU, improving the accuracy of the scoring engine. Experiments, conducted on eight real-world benchmark datasets, showed that DeLAES could understand semantics more accurately and enhance the performance of the scoring engine in terms of the quality of the final results.

Comparisons with state-of-the-art baseline methods demonstrated that DeLAES is more effective and efficient in terms of the classification quality in most of the cases. Future work will focus on integrating attention mechanism and investigating their effect on the performance of score prediction.

\begin{acknowledgements}

This research is supported by the \'UNKP-21-4 New National Excellence Program of the Ministry for Innovation and Technology from the source of the National Research, Development and Innovation Fund. Supported by the Telekom Innovation Laboratories (T-Labs), the Research and Development unit of Deutsche Telekom. Project no. ED\_18-1-2019-0030 (Application domain specific, highly reliable IT solutions subprogramme) has been implemented with the support provided from the National Research, Development and Innovation Fund of Hungary, financed under the Thematic Excellence Programme funding scheme.

\end{acknowledgements}

%
 \section*{Conflict of interest}

The authors declare that they have no conflict of interest.


\bibliographystyle{spmpsci}      
\bibliography{natbib}   

\end{document}